# Development of a Legal Document AI-Chatbot

Pranav Nataraj Devaraj
School of Computer Science and Engineering
Vellore Institute of Technology
Chennai, India
pranav.nataraj2022@vitstudent.ac.in

Rakesh Teja P V
School of Computer Science and Engineering
Vellore Institute of Technology
Chennai, India
rakeshteja.pv2022@vitstudent.ac.in

Aaryav Gangrade
School of Computer Science and Engineering
Vellore Institute of Technology
Chennai, India
aaryav.gangrade2022@vitstudent.ac.in

Manoj Kumar R
Professor
School of Computer Science and Engineering
Vellore Institute of Technology
Chennai, India

*Abstract*— With the exponential growth of digital data and the increasing complexity of legal documentation, there is a pressing need for efficient and intelligent tools to streamline the handling of legal documents. With the recent developments in the AI field, especially in chatbots, it cannot be ignored as a very compelling solution to this problem. An insight into the process of creating a Legal Documentation AI Chatbot with as many relevant features as possible within the given time frame is presented. The development of each component of the chatbot is presented in detail. Each component's workings and functionality has been discussed. Starting from the build of the Android app and the Langchain query processing code till the integration of both through a Flask backend and REST API methods.

## I. INTRODUCTION

The work was split between three members. The work was split into three components, the android application, the Langchain program and the flask app. All three were developed independently. Although the flask app required the Langchain program to be completed first in order to be fully built.

The final product we have developed is a functional and ready to deploy chatbot. It has a user interface in the form of an android application. The query processing program is written using the framework Langchain in python. The Langchain program is packaged into a flask application which contains a REST endpoint for processing queries in the form of POST requests. The flask application can then be hosted on a server and the android application communicates with the backend through POST requests. The relevant documents need to be placed inside the data directory in the backend server. The chatbot will then process the queries in the context of the documents present in the directory and provide a response.

## II. METHODOLOGY

We will now explore each component of the chatbot. We will discuss their development and the working logic behind each of them.

### A. Langchain

Langchain is an open source Natural Language Processing(NLP) framework designed to simplify the creation of applications using Large Language Models(LLM). The main purpose of using Langchain is to combine powerful LLMs with various external applications wherever there is a need for NLP. Langchain uses a feature known as Embeddings transformer where a LLM is used which in our case is OPEN AI's GPT to generate embedding related to the prompt that is given by the user. The key function of the embeddings is to understand the semantic meaning of the prompt given by the user. Once it understands the context in which the prompt was given, it looks for the relevant texts that are semantically relevant to the prompt.

When we pass a large document as an input into Langchain, it breaks the entire content into smaller chunks of text and stores them in vector store. When a prompt is given, it understands the semantic meaning and the context in which it was given and uses a technique called as "Cosine Similarity" to measure the similarity between the prompt and the information stored in the vector store. It ranks every text chunk in the vector store and retrieves the text chunk that has higher Cosine Similarity with the prompt.

Due to its capability of processing huge amount of data and responding similarly to a human, Langchain finds its application in assisting as an Interactive Chatbot, text summarization, coding assistance, marketing and e-commerce platforms to better engage with customers. In our case, it acts as the chatbot assistant to answer the legal queries of the user regarding the document which was uploaded. This can prove to be a useful tool to both the layman and the legal aspirants to understand the long and tedious judgements given by the courts by prompting questions regarding the document. The legal context is brought to the chatbot by also sending the indian constitution to the chatbot along with the document of interest.

### B. Flask Application

The technology chosen to build the backend was Flask. This component allows us to use our query processing program from anywhere by hosting it on a server. The flask app provides a POST request end point. Once the app is hosted on a server, anyone can make POST requests to it and make use of our query processing program. There are many reasons for choosing Flask as the technology to build our backend, some of which are:

- *Lightweight and Flexible Framework*: Flask is known for its simplicity and minimalism, making it easy to understand and quick to set up. This lightweight nature ensures that the chatbot application remains agile and responsive, facilitating faster development and deployment.
- *RESTful API Support*: Flask is well-suited for building RESTful APIs, and in this project, it serves as the backend with a REST endpoint for processing queries. This RESTful architecture allows for seamless communication between the android application and the backend, enhancing scalability and ease of integration with other systems if needed.
- *Python Integration*: Since the Langchain program, responsible for query processing, is written in Python, Flask, also a Python web framework, facilitates smooth integration between the android application and the processing logic. This cohesiveness simplifies the overall development process.
- *Scalability and Hosting*: Flask applications are easily deployable, and they can be hosted on various platforms, making it convenient to scale the chatbot based on demand. This feature is crucial for ensuring that the chatbot remains accessible and responsive, especially if the user base grows over time.

The code in Fig.1 initializes the flask app. It makes sure that the app is run only when the script is directly run and not just imported. It also provides the 'app.run()' method which allows us to run the development server for the app. The code also enables Cross-Origin Resource Sharing(CORS), allowing the application to handle requests from different origins.

```
app = Flask(__name__)
CORS(app)
if __name__ == "__main__":
    app.run(host="0.0.0.0", port=8095, debug=False)
```

*Fig.1 Code for initializing flask application*

The flask app provides the '/docqna' route (Fig.2) which handles POST requests. Upon an appropriate request being made to this path, the 'processclaim()' function will be executed. This function parses the POST request body as JSON then extracts the user's query from it. It then runs the function 'qa_chain()' with the query as its parameter. The 'qa_chain()' function is the query processing program. It takes the user's query as input and provides the response.

```
@app.route('/docqna',methods = ["POST"])
def processclaim():
    try:
        input_json = request.get_json(force=True)
        query = input_json["query"]
        result = qa_chain(query)
        return result['result']
    except:
        return jsonify({"Status":"Failure --- some error occured"})
```

*Fig.2 Route provided by flask application*

This path will only successfully process POST requests with a JSON body of a specific form. Even though it will try to force parse any type of content as JSON it still requires a valid 'query' object in order to process the query.

```
"query": "what does the document tell us?"
```

*Fig.3 Example of valid JSON body*

### C. Android Application

To build the android application we will be using Kotlin and XML. XML will be used for the frontend and Kotlin will be used for backend. The reasons for using the above technologies is provided below.

i)    Kotlin

• Modern Language Features: Kotlin is a modern, statically-typed programming language that brings many features to the table, making code more concise, expressive, and safer compared to Java.

• Interoperability with Java: Kotlin is fully interoperable with Java, allowing you to leverage existing Java libraries and frameworks seamlessly. This is particularly beneficial for Android development, as many Android libraries and tools are originally written in Java.

• Conciseness and Readability: Kotlin's concise syntax reduces boilerplate code, making the codebase more readable and maintainable. This can lead to increased development speed and fewer chances of introducing bugs.

• Official Language for Android: Kotlin is officially supported by Google as a first-class language for Android development. This support means that Kotlin receives regular updates, and new Android features are often Kotlin-first or Kotlin-only.

• Coroutines for Asynchronous Programming: Kotlin introduces coroutines, which simplifies asynchronous programming. This is crucial for Android apps that often involve network operations, database queries, or other tasks that should not block the main UI thread.

ii)    XML for Layouts:

• Declarative UI with XML: XML is used for defining layouts in Android, providing a declarative way to describe the UI components and their attributes. This separation of UI and logic makes it easier to understand and maintain the code.

• Resource Management: XML is used to define resources such as layouts, strings, and colors in a separate file. This allows for efficient resource management and makes it easy to support multiple device configurations.

• Data Binding: XML can be integrated with Android Data Binding, allowing for a more seamless connection between the UI components and the underlying data model. This can lead to cleaner and more maintainable code, as changes to the data automatically update the UI and vice versa.

• UI Customization and Theming: XML allows for easy customization of UI components and theming of the app. Styles and themes can be defined in XML, providing a consistent look and feel throughout the app.

• Accessibility: XML layouts can be designed with accessibility in mind, making it easier to ensure that your app is usable by people with disabilities. This includes features like content descriptions, focus order, and screen reader support.

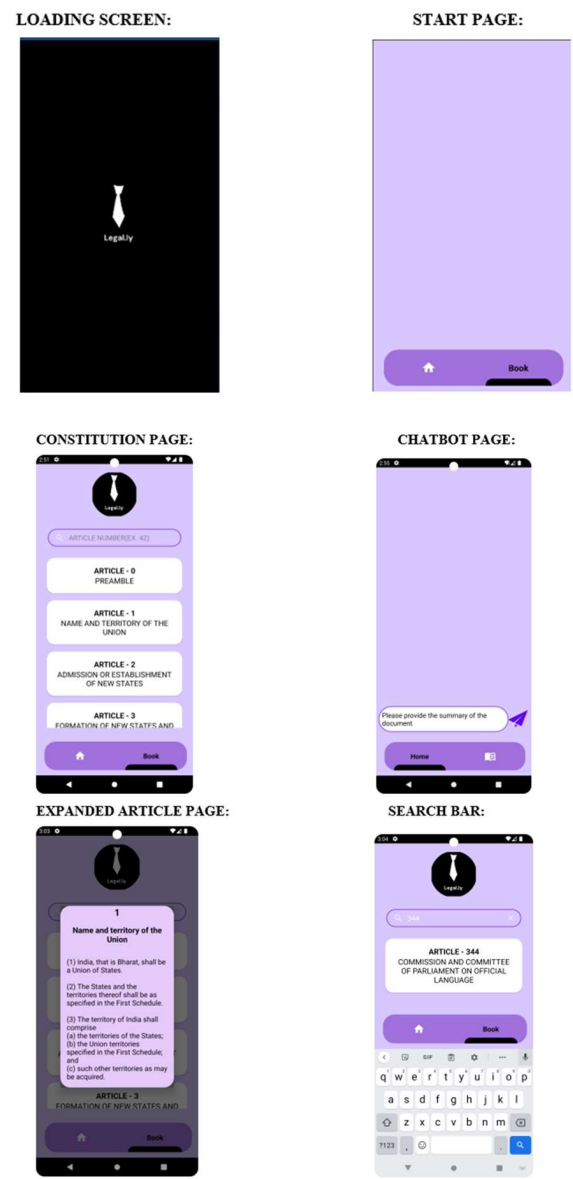

*Fig.4 The various UI components of the android application.*

In summary, choosing Kotlin for the app's logic and XML for defining layouts provides a powerful and efficient combination for Android app development, offering modern language features, seamless integration, and a declarative approach to UI design.

## III. Results

Our final product is a functional chatbot which can be used to discuss multiple documents. During development of the chatbot. Before the integration of the android application, Postman was used to test for results.

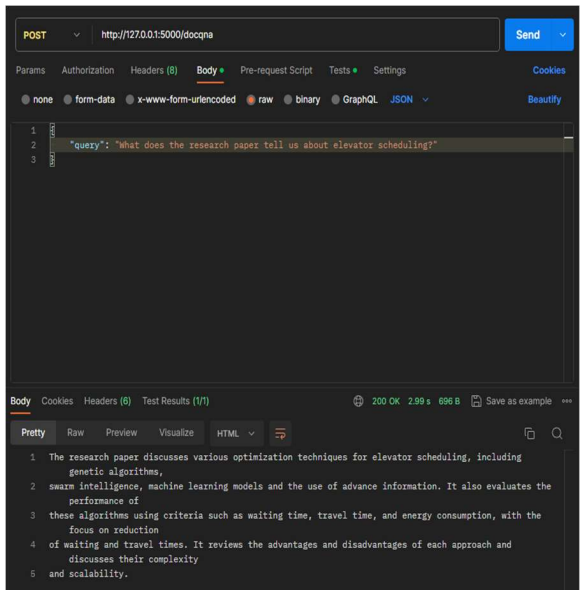

*Fig.5 Example of query and response body*

The query in the example in Fig.10 is "What does the research paper tell us about elevator scheduling?" and the response is "The research paper discusses various optimization techniques for elevator scheduling, including genetic algorithms, swarm intelligence, machine learning models and the use of advance information. It also evaluates the performance of these algorithms using criteria such as waiting time, travel time, and energy consumption, with the focus on reduction of waiting and travel times. It reviews the advantages and disadvantages of each approach and discusses their complexity and scalability."

The above query and response was asked in the context of a research paper titled "An Analysis of Various Optimization Techniques for Elevator Scheduling ". This shows that our chatbot is able to understand the document's context and respond accordingly. To turn this into a legal context chatbot we feed the Indian constitution into the bot long with the document of interest.

## IV. Literature Review

The article [1] serves as a guide to beginners to learn about Langchain. It clearly explains the process of installing Langchain in the system using specific commands. It briefly covers about the process in which Langchain operates starting from getting the document from the user as input to breaking it into smaller text chunks and finding out their semantic meaning using the Cosine Similarity technique and then getting the query from user and finding and fetching the response from the Large Language Module. It also explains the intricate processes such as Embeddings transformation and vector storage of the text chunks in a plain and simple language. The article also explains about the procedure to interact with multiple documents at the same time by creating a List of Documents.

The part 1 of the article [2] provides key insights about the different modules and the various different features that each module has to offer. It explains about 7 important modules including their functionalities and the way in which Langchain chains all of these modules into its framework. The article also provides code snippets to implement these functionalities. The part 2 of the article explains the way in which Langchain processes the data using Embeddings and Vector store. It also covers the key details about the function which is used to remember the previous conversions and correlate with the queries.

[4] is the official documentation of Kotlin which was used for learning the syntax of Kotlin. [5] is the course by Google which was used to understand and implement Kotlin. The video [6] was used  to understand XML and how to implement XML and Kotlin together.

The blog [3] by Postman was used to understand the basics of API testing. The tutorial video [7] was used to understand how to make a Flask application and make use of REST APIs in Flask. The documentation of Flask [8] was also used as a reference material.

## V. Conclusion

In conclusion, this project followed an efficient methodology in order to build a functional and scalable chatbot. Future work on this project holds great potential in turning this into an industry standard chatbot. There are many features such as AI training, provision in android app for uploading documents and increasing query token limit which can be developed in the future. Work can also be done on the UI to improve the user experience. Overall, a legal document chatbot able to answer questions within the context of Indian constitution was developed.